\documentclass[twoside,11pt]{article}

\usepackage{blindtext}

\usepackage[nohyperref,preprint]{jmlr2e}

\usepackage{amsmath}		\usepackage{amssymb}		\usepackage{amsfonts}		

\usepackage{mathtools}		

\mathtoolsset{}

\usepackage[utf8]{inputenc}		\usepackage[T1]{fontenc}

\usepackage{dsfont}

\usepackage{acronym}

\usepackage[labelfont={bf,small},labelsep=colon,font=small]{caption}	

\usepackage[dvipsnames,svgnames]{xcolor}		\colorlet{MyRed}{Crimson!75!black}
\colorlet{MyGreen}{DarkGreen!80!black}
\colorlet{MyBlue}{MediumBlue!80!black}
\definecolor{ermtrain}{rgb}{0., 0.5, .9}
\definecolor{ermtest}{rgb}{0., 1., 0.}
\definecolor{robusttrain}{rgb}{1., 0., 0.}

\usepackage{cancel}		\usepackage{latexsym}		

\usepackage{pifont}

\usepackage{subcaption}		\usepackage{tikz}		\usetikzlibrary{calc,patterns}		\usepackage{pgfplots}
\usepgfplotslibrary{groupplots,dateplot}
\usetikzlibrary{patterns,shapes.arrows}
\pgfplotsset{compat=newest}

\usepackage{array}		\usepackage{booktabs}		\usepackage[inline,shortlabels]{enumitem}		\setenumerate{itemsep=\smallskipamount,topsep=\medskipamount}

\usepackage{xspace}

\usepackage{hyperref}
\hypersetup{
colorlinks=false,
linktocpage=true,
pdfstartview=FitH,
breaklinks=true,
pdfpagemode=UseNone,
pageanchor=true,
pdfpagemode=UseOutlines,
plainpages=false,
bookmarksnumbered,
bookmarksopen=false,
bookmarksopenlevel=1,
hypertexnames=true,
pdfhighlight=/O,
urlcolor=MyBlue,linkcolor=MyBlue,citecolor=MyBlue,	pdftitle={},
pdfauthor={},
pdfsubject={},
pdfkeywords={},
pdfcreator={pdfLaTeX},
pdfproducer={LaTeX with hyperref}
}

\usepackage[sort&compress,capitalize,nameinlink]{cleveref}		\crefname{algo}{Algorithm}{Algorithms}
\crefname{assumption}{Assumption}{Assumptions}
		\crefformat{equation}{(#2#1#3)}
\crefmultiformat{equation}{(#2#1#3)}{ and~(#2#1#3)}{, (#2#1#3)}{ and~(#2#1#3)}
\crefrangeformat{equation}{\upshape(#3#1#4)\textendash(#5#2#6)}

\usepackage{autonum}

\usepackage{listings}

\definecolor{codegreen}{rgb}{0,0.6,0}
\definecolor{codegray}{rgb}{0.5,0.5,0.5}
\definecolor{codepurple}{rgb}{0.58,0,0.82}
\definecolor{backcolour}{rgb}{0.95,0.95,0.92}

\lstdefinestyle{mystyle}{
    backgroundcolor=\color{backcolour},   
    commentstyle=\color{codegreen},
    keywordstyle=\color{magenta},
    numberstyle=\tiny\color{codegray},
    stringstyle=\color{codepurple},
    basicstyle=\ttfamily\footnotesize,
    breakatwhitespace=false,         
    breaklines=true,                 
    captionpos=b,                    
    keepspaces=true,                 
    numbers=left,                    
    numbersep=5pt,                  
    showspaces=false,                
    showstringspaces=false,
    showtabs=false,                  
    tabsize=2
}

\usepackage{algorithm}		\usepackage{algpseudocode}

\usepackage{thmtools}		\usepackage{thm-restate}

\makeatletter
\def\cleartheorem#1{\expandafter\let\csname#1\endcsname\relax
    \expandafter\let\csname c@#1\endcsname\relax
}
\makeatother
\cleartheorem{example}
\cleartheorem{theorem}
\cleartheorem{lemma}
\cleartheorem{proposition}
\cleartheorem{definition}
\cleartheorem{corollary}
\cleartheorem{remark}

\newcounter{proofpart}

\numberwithin{example}{section}

\usepackage[sort&compress,capitalize,nameinlink]{cleveref}		\crefname{algo}{Algorithm}{Algorithms}
\crefname{assumption}{Assumption}{Assumptions}
\crefname{lemma}{Lemma}{Lemmas}
		\crefformat{equation}{(#2#1#3)}
\crefmultiformat{equation}{(#2#1#3)}{ and~(#2#1#3)}{, (#2#1#3)}{ and~(#2#1#3)}
\crefrangeformat{equation}{\upshape(#3#1#4)\textendash(#5#2#6)}

\usepackage{autonum}

\usepackage[normalem]{ulem}

\newcommand{\debug}[1]{#1}

\newcommand{\newmacro}[2]{\newcommand{#1}{{\debug{#2}}}}		\newcommand{\newop}[2]{\DeclareMathOperator{#1}{\debug{#2}}}

		\DeclarePairedDelimiter{\bracks}{[}{]}

\DeclarePairedDelimiterX{\setdef}[2]{\{}{\}}{#1:#2}		\DeclarePairedDelimiterXPP{\exclude}[1]{\mathopen{}\setminus}{\{}{\}}{}{#1}

								\newcommand{\R}{\mathbb{R}}

								\DeclarePairedDelimiterXPP{\bigof}[1]{\mathcal{O}}{(}{)}{}{#1}												\DeclareMathOperator{\dist}{dist}																		\DeclareMathOperator{\one}{\mathds{1}}								\DeclareMathOperator{\relint}{ri}

		\newcommand{\eg}{e.g.,\xspace}

\newcommand{\alt}[1]{#1'}		\newcommand{\altalt}[1]{#1''}				

\newmacro{\dd}{\mathrm{d}}						\newcommand{\eps}{\varepsilon}

\newmacro{\const}{c}		\newmacro{\Const}{C}		\usepackage{xparse}
\NewDocumentCommand{\coef}{O{\lambda}}{\debug{#1}}
\newmacro{\param}{\theta}		\newmacro{\params}{\Theta}

\newmacro{\beforestart}{0}		\newmacro{\start}{1}		\newmacro{\afterstart}{2}		\newmacro{\running}{\start,\afterstart,\dotsc}		\newmacro{\halfrunning}{1,3/2,2\dotsc}		

\newmacro{\run}{t}		\newmacro{\runalt}{s}		\newmacro{\runaltalt}{\tau}		\newmacro{\nRuns}{T}		\newmacro{\runs}{\mathcal{\nRuns}}

\newmacro{\state}{X}		\newmacro{\statealt}{Y}		\newmacro{\statealtalt}{Z}

\newmacro{\tstart}{0}				\newmacro{\timealt}{s}		\newmacro{\horizon}{T}		

\newmacro{\traj}{x}		\newmacro{\trajalt}{y}		\newmacro{\trajaltalt}{z}		

\newmacro{\flowmap}{\Theta}		\DeclarePairedDelimiterXPP{\flowof}[2]{\flowmap_{#1}}{(}{)}{}{#2}

\newop{\Nash}{NE}		\newop{\CE}{CE}		\newop{\CCE}{CCE}		\newop{\NI}{NI}		

\newop{\brep}{br}		\newop{\preg}{\overline{Reg}}		\newop{\val}{val}

\newmacro{\play}{i}		\newmacro{\playalt}{j}		\newmacro{\playaltlalt}{k}		\newmacro{\nPlayers}{N}		\newmacro{\players}{\mathcal{\nPlayers}}		

\newmacro{\pure}{\alpha}		\newmacro{\purealt}{\beta}		\newmacro{\purealtalt}{\gamma}		\newmacro{\nPures}{A}		\newmacro{\pures}{\mathcal{\nPures}}		

\newmacro{\loss}{\ell}		\newmacro{\pay}{u}		\newmacro{\payv}{v}		\newmacro{\pot}{f}		

\newmacro{\game}{\mathcal{G}}		\newmacro{\gamefull}{\game(\players,\points,\pay)}		

\newmacro{\fingame}{\Gamma}		\newmacro{\fingamefull}{\Gamma(\players,\pures,\pay)}

\newmacro{\gmat}{g}		\newmacro{\gdist}{\dist_{\gmat}}
\newmacro{\mfld}{M}		\newmacro{\form}{\omega}		

\newmacro{\tvec}{z}		\newmacro{\uvec}{u}		

\newmacro{\ball}{\overline{\mathbb{B}}}		\newmacro{\openball}{\mathbb{B}}		\newmacro{\sphere}{\mathbb{S}}

\newmacro{\graph}{\mathcal{G}}
\newmacro{\vertices}{\mathcal{V}}
\newmacro{\edges}{\mathcal{E}}

\newmacro{\mat}{M}		\newmacro{\hmat}{H}		

\newmacro{\ones}{\mathbf{1}}		\newmacro{\eye}{I}		\newmacro{\zer}{\mathbf{0}}

		\DeclarePairedDelimiterXPP{\dnorm}[1]{}{\lVert}{\rVert}{_{\ast}}{#1}		

		\DeclarePairedDelimiterXPP{\onenorm}[1]{}{\lVert}{\rVert}{_{1}}{#1}		\DeclarePairedDelimiterXPP{\twonorm}[1]{}{\lVert}{\rVert}{_{2}}{#1}		\DeclarePairedDelimiterXPP{\supnorm}[1]{}{\lVert}{\rVert}{_{\infty}}{#1}

				\DeclarePairedDelimiterX{\braket}[2]{\langle}{\rangle}{#1,#2}		\DeclarePairedDelimiterX{\inner}[1]{\langle}{\rangle}{#1}

\newmacro{\vecspace}{\mathcal{V}}		\newmacro{\subspace}{\mathcal{W}}		

\newmacro{\coord}{i}		\newmacro{\coordalt}{j}		\newmacro{\coordaltalt}{k}		\newmacro{\nCoords}{d}		\newmacro{\dims}{\nCoords}		\newmacro{\vdim}{\nCoords}		

\newmacro{\pvec}{v}		

\newmacro{\bvec}{e}		\newmacro{\bvecs}{\mathcal{E}}

\newmacro{\pspace}{\mathcal{X}}		\newmacro{\dspace}{\pspace^{\ast}}		

\newmacro{\dvec}{\dpoint}		\newmacro{\dbvec}{\eps}		

\newmacro{\dpoint}{y}		\newmacro{\dpointalt}{\alt\dpoint}		\newmacro{\dpointaltalt}{\altalt\dpoint}		\newmacro{\dpoints}{\mathcal{Y}}		

\newmacro{\dstate}{Y}		\newmacro{\dbase}{w}

\newmacro{\dualvar}{\lambda}
\newmacro{\dualvaralt}{\mu}
\newmacro{\dualfunc}{\phi}
\newmacro{\dualfuncalt}{\psi_\radius}
\newmacro{\empdualfunc}{\phi_\nsamples}
\newmacro{\lbdualvar}{\lb{\dualvar}}
\newmacro{\ubdualvar}{\ub{\dualvar}}
\newmacro{\basedualvar}{\dualvar^*_0}

\newop{\Opt}{Opt}		\newop{\Sol}{Sol}		\newop{\gap}{Gap}		\newop{\orcl}{Or}		

\newop{\primal}{(P)}
\newop{\dual}{(Q)}

\newmacro{\tfun}{g}		\newmacro{\obj}{f}		\newmacro{\objalt}{g}		\newmacro{\objaltalt}{h}		\newmacro{\sobj}{F}		

\newmacro{\gvec}{g}		\newmacro{\oper}{A}		\newmacro{\vecfield}{v}		

\newcommand{\sol}[1][\point]{{#1}^{\star}}				\newmacro{\vecsol}{\vecfield(\sol)}

\newmacro{\signal}{V}		\newmacro{\step}{\gamma}		\newmacro{\learn}{\eta}		

\newmacro{\vbound}{G}		

\newmacro{\lips}{Lip}		\newmacro{\lipcst}{L}
\newmacro{\bdcst}{M}

\newmacro{\bdcstalt}{\widetilde{F}}
\newmacro{\fbound}{\bdcstalt(0)}		

\newmacro{\varbdcst}{K}
\newmacro{\lbcst}{a}
\newmacro{\ubcst}{b}
\newmacro{\strong}{\mu}		\newmacro{\strongOpt}{\mu^{*}}		\newmacro{\smooth}{L_2}
\newmacro{\hsmooth}{L_3}
\newmacro{\tmplipcst}{L}
\newmacro{\tmpbdcst}{B}

\newop{\cone}{cone}
\newop{\tspace}{T}		\newop{\tcone}{TC}		\newop{\dcone}{DC}		\newop{\ncone}{NC}		\newop{\regncone}{\widehat{NC}}		\newop{\pcone}{PC}		\newop{\hull}{\Delta}		

\newmacro{\cvx}{\mathcal{C}}		\newmacro{\subd}{\partial}

\newmacro{\minmax}{\mathcal{L}}		

\newmacro{\minvar}{\point_{1}}		\newmacro{\minvaralt}{\alt\minvar}		\newmacro{\minvars}{\points_{1}}		\newmacro{\minsol}{\sol[\minvar]}		

\newmacro{\maxvar}{\point_{2}}		\newmacro{\maxvaaltr}{\alt\maxvar}		\newmacro{\maxvars}{\points_{2}}		\newmacro{\maxsol}{\sol[\maxvar]}

\newop{\Eucl}{\Pi}		\newop{\logit}{\Lambda}		\newop{\dkl}{KL}		

\newmacro{\hreg}{h}		\newmacro{\hconj}{\hreg^{\ast}}		\newmacro{\breg}{D}		\newmacro{\mprox}{P}		\newmacro{\mirror}{Q}		\newmacro{\fench}{F}		\newmacro{\hstr}{K}		\newmacro{\depth}{H}		\newmacro{\proxdom}{\points_{\hreg}}		\newmacro{\zone}{\mathbb{D}}		\newmacro{\bregkernel}{\theta} 

\DeclarePairedDelimiterXPP{\proxof}[2]{\mprox_{#1}}{(}{)}{}{#2}

\newmacro{\bregexp}{\alpha}
\newmacro{\bregcst}{M}

\newmacro{\kernelcst}{C}
\newmacro{\kernelexp}{q}

\newmacro{\point}{x}		\newmacro{\pointalt}{\alt\point}		\newmacro{\pointaltalt}{\altalt\point}		\newmacro{\points}{\mathcal{K}}		\newmacro{\intpoints}{\relint\points}		

\newmacro{\basealt}{q}		\newmacro{\basealtalt}{u}		

\newmacro{\open}{\mathcal{U}}		\newmacro{\closed}{\mathcal{C}}		\newmacro{\cpt}{\mathcal{K}}		\newmacro{\nbd}{\mathcal{U}}		\newmacro{\nhd}{\mathcal{U}}		\newmacro{\mset}{A}

\NewDocumentCommand{\ex}{E{_}{\coupling} o}{\mathbb{E}_{#1}\IfValueT{#2}{\bracks*{#2}}}		

\newop{\prob}{P}\newcommand{\dirac}[1]{\debug{\delta}_{#1}}

\newop{\empirical}{\prob_{\nsamples}}\newop{\emp}{\empirical}\newop{\proba}{\mathbb{P}}\newmacro{\empex}{\frac{1}{\nsamples}\sum_{\ind=1}^\nsamples}
\newop{\probalt}{Q}
\newop{\coupling}{\pi}
\newop{\couplingalt}{\alt\pi}
\NewDocumentCommand{\couplings}{e{_} O{\samples}}{\debug{\mathcal{P}}\IfValueT{#1}{_{#1}}(#2\times#2)}
\newop{\Var}{Var}		\newop{\simplex}{\hull}		

\newop{\rad}{s}

\DeclarePairedDelimiterXPP{\exof}[1]{\ex}{[}{]}{}{ #1}

\DeclarePairedDelimiterXPP{\probof}[2]{#1}{(}{)}{}{ #2}

\DeclarePairedDelimiterXPP{\oneof}[1]{\one}{\{}{\}}{}{ #1}

\newmacro{\sample}{\xi}		\newmacro{\samplealt}{\altsample} \newmacro{\altsample}{\zeta}
\newmacro{\samplealtalt}{\alt\altsample}
\newmacro{\samples}{\Xi}		

\newmacro{\optsample}{\xi^*}
\newmacro{\optoptsample}{\xi^{*}}

\newmacro{\nsamples}{n}
\newmacro{\Nsamples}{N}
\newmacro{\nsamplesalt}{m}

\newmacro{\target}{y}
\newmacro{\Point}{X}
\newmacro{\Target}{Y}
\newmacro{\Pointalt}{\alt\Point}
\newmacro{\Targetalt}{\alt\Target}

\newmacro{\targetalt}{\alt\target}
\newmacro{\targetcost}{\kappa}

\newmacro{\distance}{d}
\newmacro{\distfunc}{D}
\newmacro{\Hdist}{D_H}

\newmacro{\cost}{c} 
\newmacro{\Cost}{C}
\newmacro{\mincost}{\sol[c]}
\newmacro{\maxcost}{\sol[C]}

\NewDocumentCommand{\wass}{s e{^} O{} o m}{\debug{W}\IfValueT{#2}{^{{#2}}}\IfValueT{#4}{^{#4}}_{#3}\IfBooleanT{#1}{\left}(#5\IfBooleanT{#1}{\right})}

\newmacro{\filter}{\mathcal{F}}		\newmacro{\probspace}{(\samples,\filter,\prob)}		

\newmacro{\radius}{\rho}
\newmacro{\minradius}{{\rho^2_{\nsamples}}}
\newmacro{\cstminradius}{{\rho_{min}}}

\newmacro{\GeoRad}{R}
\newmacro{\OptGeoRad}{R^*}
\newmacro{\georad}{r}

\newmacro{\Margin}{\Delta}

		\newmacro{\event}{E}       \newmacro{\eventalt}{H}       \newmacro{\mean}{\mu}		\newmacro{\sdev}{\sigma}		\newmacro{\variance}{\sdev^{2}}		

\newmacro{\level}{\alpha}
\newmacro{\levelalt}{p}

\NewDocumentCommand{\Lspace}{E{^}{{1}} O{\prob}}{\debug{L^{#1}(#2)}}

\NewDocumentCommand{\Obj}{e{^} E{_}{\radius} D(){\obj, \prob}}{
    {F}\IfValueT{#1}{^{#1}}\IfValueT{#2}{_{#2}}(#3)
}

\NewDocumentCommand{\Objalt}{e{^} E{_}{\radius} D(){\obj}}{
    {G}\IfValueT{#1}{^{#1}}\IfValueT{#2}{_{#2}}(#3)
}

\newmacro{\ind}{i}
\newmacro{\indalt}{j}

\newmacro{\reg}{\varepsilon}
\newmacro{\regalt}{\delta}
\newmacro{\basereg}{\base[\reg]}
\newmacro{\basesdev}{\base[\sdev]}
\newmacro{\regparam}{\varepsilon}
\newmacro{\regparamalt}{\delta}
\newmacro{\regparamaltalt}{\tau}
\newmacro{\Regparam}{\Delta}

\newmacro{\rv}{X}
\newmacro{\altrv}{Y}
\newmacro{\discrprob}{p}
\newmacro{\discrprobalt}{q}

\NewDocumentCommand{\measures}{O{\samples^2} e{^}}{\mathcal{M}\IfValueT{#2}{^{#2}}(#1)}

\newcommand{\base}[1][\coupling]{{#1}_0}

\newmacro{\scalar}{u}

\newmacro{\exponent}{p}
\newmacro{\pexp}{p}
\newmacro{\qexp}{q}

\newmacro{\volcst}{V}

\newmacro{\funcs}{\mathcal{F}}

\newmacro{\func}{\obj}
\newmacro{\funcalt}{\objalt}
\newmacro{\funcaltalt}{\objaltalt}

\newmacro{\proper}{\tau}				

\newmacro{\error}{Z}		\newmacro{\bias}{b}		\newmacro{\brown}{W}		

\newmacro{\serror}{\theta}		\newmacro{\snoise}{\xi}		\newmacro{\sbias}{\psi}		

\newmacro{\sbound}{M}		\newmacro{\bbound}{B}		\newmacro{\noisepar}{\sdev}		\newmacro{\noisevar}{\variance}		

\newcounter{cnstcnt}

\newmacro{\uncertainty}{U}

\NewDocumentCommand{\risk}{oo}{
    \boldsymbol{\mathcal{R}}\IfValueT{#2}{^{#2}}\IfValueT{#1}{_{#1}}(\obj)
}

\NewDocumentCommand{\emprisk}{oo}{
    \widehat{\mathcal{R}}\IfValueT{#2}{^{#2}}\IfValueT{#1}{_{#1}}(\obj)
}

\NewDocumentCommand{\expect}{om}{\mathbb{E}\IfValueT{#1}{_{#1}}{\bracks*{#2}}}		\NewDocumentCommand{\probab}{om}{\mathbb{P}\IfValueT{#1}{_{#1}}{\bracks*{#2}}}

\newacro{LHS}{left-hand side}
\newacro{RHS}{right-hand side}
\newacro{iid}[i.i.d.]{independent and identically distributed}
\newacro{lsc}[l.s.c.]{lower semi-continuous}
\newacro{usc}[u.s.c.]{upper semi-continuous}
\newacro{rv}[r.v.]{random variable}
\newacro{NE}{Nash equilibrium}

\newacro{ABP}{abstract Bregman proximal}
\newacro{BP}{Bregman proximal}

\newacro{DGF}{distance-generating function}
\newacro{EG}{extra-gradient}
\newacro{MP}{mirror-prox}
\newacro{MD}{mirror descent}
\newacro{OMD}{optimistic mirror descent}
\newacro{OMWU}{optimistic multiplicative weights update}
\newacro{PMP}{past mirror-prox}
\newacro{AMP}{abstract mirror-prox}
\newacro{MPT}{mirror-prox template}

\newacro{VI}{variational inequality}
\newacro{VIP}{variational inequality problem}
\newacro{KKT}{Karush\textendash Kuhn\textendash Tucker}
\newacro{FOS}{first-order stationary}
\newacro{SOO}{second-order optimality}
\newacro{SOS}{second-order sufficient}
\newacro{DGF}{distance-generating function}
\newacro{SFO}{stochastic first-order oracle}

\newacro{DRO}{distributionally robust optimization}
\newacro{WDRO}{Wasserstein distributionally robust optimization}
\newacro{ML}{machine learning}
\newacro{SVM}{support vector machines}
\newacro{ERM}{empirical risk minimization}
\newacro{OT}{optimal transport}
\newacro{ELBO}{evidence lower bound}
\newacro{MCMC}{Monte Carlo Markov Chain}
\newacro{SAEM}{stochastic approximation expectation-maximization}
\newacro{AD}{automatic differentiation}
\newacro{OR}{operational research}
\newacro{PAC}{probably approximately correct}
\newacro{SA}{stochastic approximation}

\newacro{KL}{Kullback-Leibler}

\usepackage{lastpage}

\ShortHeadings{\texttt{skwdro}: a library for Wasserstein distributionally robust machine learning}{Vincent, Azizian, Malick, and Iutzeler}
\firstpageno{1}

\begin{document}

\title{\texttt{skwdro}: a library for\\ Wasserstein distributionally robust machine learning}

\author{\name Florian Vincent, Waïss Azizian, Jérôme Malick\\ 
\email firstname.name@univ-grenoble-alpes.fr \\
       \addr Univ. Grenoble Alpes, CNRS, Grenoble INP, LJK, F-38000 Grenoble, France\\
       \AND
       \name Franck Iutzeler \email franck.iutzeler@math.univ-toulouse.fr \\
       \addr Université de Toulouse, CNRS, UPS, F-31062 Toulouse, France}
\editor{Alexandre Gramfort}

\maketitle

\begin{abstract}

We present \texttt{skwdro}, a Python library for training robust machine learning models. The library is based on distributionally robust optimization using {Wasserstein distances, popular in optimal transport and machine learning}s.
{The goal of the library is to make the training of robust models easier for a wide audience by proposing a wrapper for PyTorch modules, enabling model loss' robustification with minimal code changes.
It comes along with \texttt{scikit-learn} compatible estimators for some popular objectives.}
The core of the implementation relies on an entropic smoothing of the original robust objective, in order to ensure maximal model flexibility. 
The library is available at \url{https://github.com/iutzeler/skwdro} and the documentation at \url{https://skwdro.readthedocs.io}.

\end{abstract}


\section{Introduction: ERM, WDRO, and entropic regularization}

Training machine learning models typically relies on \ac{ERM}, which consists in minimizing the expectation of
the loss of a model on
the empirical distribution of training data.
This approach has been questioned for its limited resilience and reliability, as machine learning systems have become widely used and deployed. Recently, \ac{DRO} has emerged as a powerful paradigm towards better generalization and better resilience against data heterogeneity and distribution shifts; see \citet{chen2020distributionally}, \cite{blanchet2021statistical}, and \cite{kuhn2025distributionally}.

The paradigm consists in minimizing the \emph{average loss} over the \emph{worst distribution} in a neighborhood of the empirical distribution, thus capturing any data uncertainty. A natural way to define this neighborhood is through the Wasserstein distance (see \eg the textbook\;\cite{peyre2019computational}), as proposed by \citep{esfahani2018data}. This framework, called \ac{WDRO}, is attractive, since it combines powerful modeling capabilities, strong generalization properties, and practical robustness guarantees; 
see \eg recent theoretical and practical developments \citep{kuhn2019wasserstein, azizian2024exact,blanchet2024distributionally}. Unfortunately, the resulting optimization problem is tractable {only in specific cases} (e.g.\;\cite{esfahani2018data}, \cite{belbasi2023s}), which limits the use of \ac{WDRO} in practice for training robust models. 

In this work, we tackle this computational issue and provide a complete and easy-to-use library for numerical \ac{WDRO}. In \cref{sec:models}, we formalize the notation and sketch our approach. We present the library \texttt{skwdro} in \cref{sec:library} and mention the numerical ingredients in \cref{sec:issues}.  The online documentation completes this short article by presenting detailed features,{tutorials}, and numerical illustrations.

\section{WDRO models and entropic variants}\label{sec:models}

Formally, we aim at selecting a model among a family parametrized by $\param \in \params$, whose error is measured by some loss function $\obj_\param(\sample)$.
Given $\nsamples$ data points $(\sample_\ind)_{\ind=1}^\nsamples$ (raw data or input-label pairs), the \ac{ERM} problem writes
\begin{equation}
        \min_{\param \in \params} ~~\empex \obj_\param(\sample_\ind) = \ex_{\sample \sim \empirical} [\obj_\param(\sample)] 
        \quad\text{ where the empirical distribution is } 
        \empirical = \empex \dirac{\sample_\ind}.
        \label{eq:ERM}
\end{equation}
As mentioned in introduction, this approach can fail if the number of data points is too small or the data distribution at deployment differs from the training distribution.
\ac{WDRO} is a way to remedy these issues: it consists in minimizing the \emph{worst expectation of the loss} when the probability distribution $Q$ lives in a \emph{Wasserstein ball} at the empirical distribution\;$\empirical$. Thus, for a robustness radius $\radius\geq 0$, this leads to the \ac{WDRO} problem
\begin{equation}\label{eq:intro_dro}\min_{\param \in \params} ~ \sup_{\wass[\cost]{\empirical, \probalt} \leq \radius}\ex_{\sample \sim \probalt}[\obj_\param(\sample)]\, ,
\end{equation}
where $\wass[\cost]{\empirical,\probalt}$ denotes
the {optimal transport cost} between $\empirical$ and $\probalt$ with ground cost $\cost : \samples\times\samples \to \R_+$, often referred to as the Wasserstein distance by a slight abuse of terminology\footnote{{The $p$-Wasserstein distance corresponds to the optimal transport cost when the ground cost is a distance between the two inputs to some power $p \geq 1$ (e.g. $c(\xi, \zeta) = \|\xi -\zeta\|^p$). Notice that our library supports any norm $\| \cdot \|$ and any power $p$ (excepted the special case $p=\infty$), as well any user-provided\;ground\;cost. }}. Applying the Lagrangian duality gives the following dual problem
\begin{align}
 \min_{\param\in\params} ~ \min_{\dualvar \geq 0} ~ \dualvar \radius +  \ex_{\sample \sim \empirical} \left[ \sup_\altsample ~\left\{\obj_\param(\altsample) - \dualvar \cost(\sample, \altsample)\right\} \right]
        \label{eq:dualWDRO}
\end{align}
where $\dualvar \geq 0$ is the dual variable associated to the Wasserstein distance constraint. For some specific tasks, this problem can be reformulated as  a convex program \citep{kuhn2019wasserstein,shafieezadehabadehregularizationmasstransportation2019}. However, this convex approach does not allow for complex models, such as neural networks. Thus, though very attractive, this robust approach usually leads to an intractable optimization problem.

To remedy this issue, we consider the following smoothed counterpart of \eqref{eq:dualWDRO}
\begin{align}
 \min_{\param\in\params} ~ \min_{\dualvar \geq 0} ~ \dualvar \radius +  \regparam ~ \ex_{\sample \sim \empirical} \left[ \log\left(\ex_{\altsample \sim \mathcal{N}(\sample, \sigma^2 I)}\left[ \exp{\left(\frac{\obj_\param(\altsample) - \dualvar \cost(\sample, \altsample)}{\regparam }\right)}\right]\right) \right]
        \label{eq:EntWDRO}
\end{align}
where the regularization strength $\regparam$ and the sampling variance $\sigma^2$ are the two parameters controlling the approximation. 
This problem interprets as the dual of the entropy regularized \ac{WDRO} problem\;\citep{azizian2022regularization, wang2021sinkhorn}. Interestingly, such entropic-regularized problems have been proved to retain the generalization guarantees of the original robust problem\;\eqref{eq:dualWDRO} (see \eg \citet{azizian2024exact,le2025universal}). {They have also opened up interesting convergence analysis (see the recent \citep{le2025unregularized})}.

In our work, we leverage the practical interest of the approach: \eqref{eq:EntWDRO} can be solved using (stochastic) gradient-based methods, which thus significantly broadens the applicability and numerical tractability of distributionally robust machine learning {based on WDRO.} This is at the core of \texttt{skwdro}.

\section{\texttt{skwdro}: implementation and features}\label{sec:library}

\texttt{skwdro} is an open-source Python library (\url{https://github.com/iutzeler/skwdro}) for machine learning with \ac{WDRO} based on the methodology presented in the previous section. 

\begin{figure}[h!]
\centering
\vspace*{3ex}
    \includegraphics[width=0.67\linewidth,trim={0cm 3.5cm 0 3cm}]{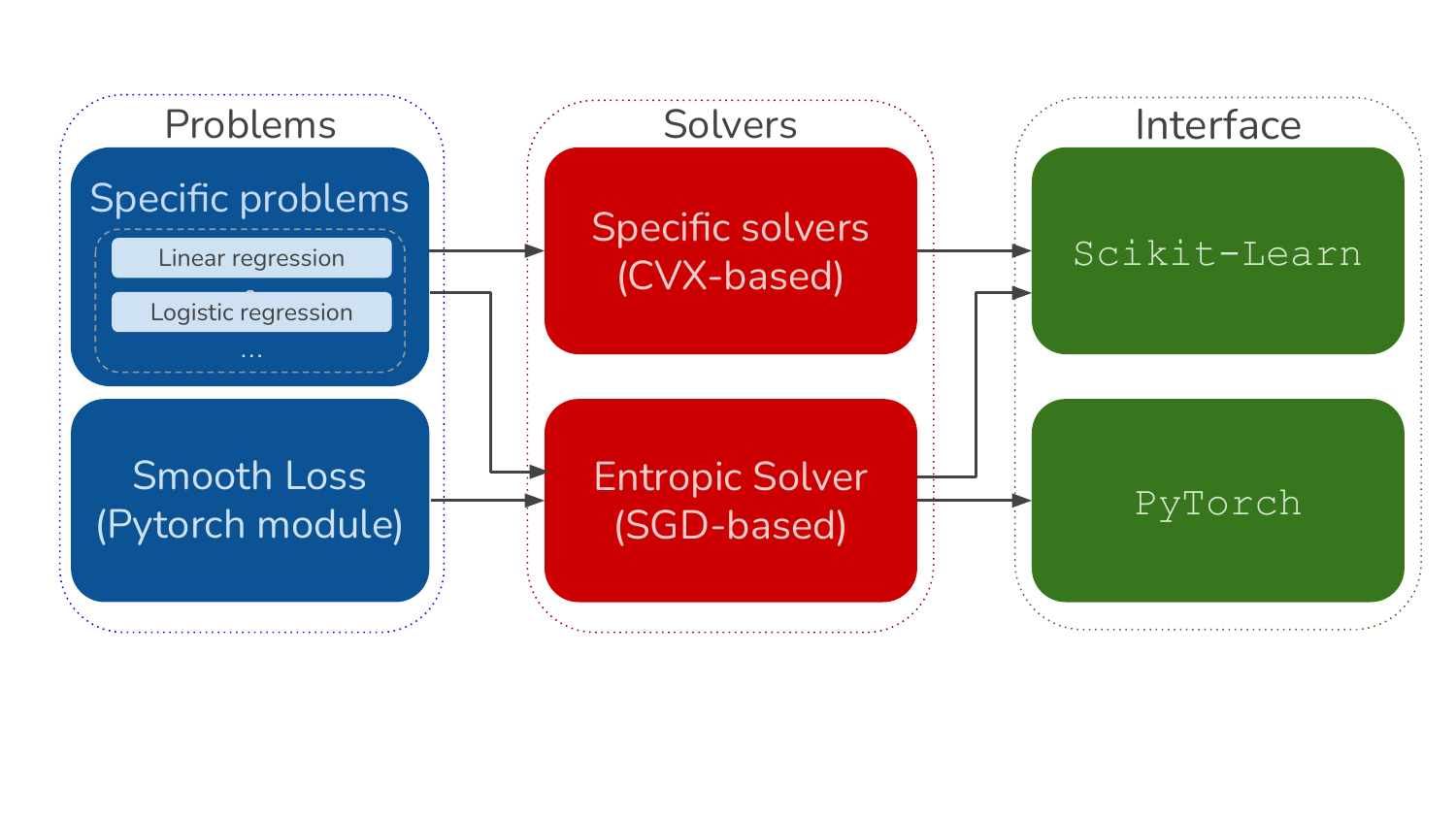}
    \caption{Schematic view of the main blocks of \texttt{skwdro}.\label{fig:skWDRO}}
\end{figure}

{The library provides an easy-to-use \texttt{PyTorch} interface for making arbitrary learning models more robust: \texttt{skwdro} first formulates the robust training of the model as an optimization problem of the form \eqref{eq:EntWDRO}; then the optimization follows the usual pipeline (e.g. using Adam or other stochastic gradient-based optimizers). The key numerical ingredients to handle the robust objective function are mentioned in \cref{sec:issues} and are further detailed in the online documentation (\url{https://skwdro.readthedocs.io/}).

Furthermore, for simpler problems (e.g.\;linear models with $c(\xi,\zeta)=\|\xi-\zeta\|^2_2$) where \eqref{eq:intro_dro} can be reformulated as a standard convex problem, \texttt{skwdro} also proposes a \texttt{scikit-learn} interface \citep{pedregosa2011scikit}, implementing known techniques (e.g.\;\cite{kuhn2019wasserstein}), and relying on \texttt{cvxpy} for optimization \citep{diamond2016cvxpy}. However this is not the main focus of \texttt{skwdro}, and, in this case, we refer users to the library \texttt{python-dro} (\cite{liu2025dro}, \url{https://python-dro.org}) to experiment with other uncertainty neighborhoods beyond Wasserstein. To summarize, \cref{fig:skWDRO} gives a schematic view of \texttt{skwdro}.}

The \texttt{PyTorch} interface of \texttt{skwdro} allows to robustify machine learning models, specified as \texttt{PyTorch} modules, with minimal code changes.  We display below a simple color-coded example: in \textcolor{Navy}{blue}, the code for learning a linear classifier with MSE loss; in \textcolor{ForestGreen}{green} (resp.  \textcolor{BrickRed}{red}), the lines to \textcolor{ForestGreen}{add} (resp. \textcolor{BrickRed}{remove}) in order to robustify the model.

\lstset{
    escapeinside={(*@}{@*)},
    commentstyle=\color{gray},
    basicstyle=\color{Navy}\ttfamily\footnotesize
}

\begin{lstlisting}[language=Python,
]
import torch as pt
import torch.nn as nn
import torch.optim as optim
from torch.utils.data import DataLoader, TensorDataset
(*@\textcolor{ForestGreen}{from skwdro.torch import robustify}@*)
# Toy data
n_features = 3
X = pt.randn(32, n_features)
y = X @ pt.rand(n_features, 1) + 1.
train_loader = DataLoader(TensorDataset(X, y), batch_size=4)
# Define the model
model = nn.Linear(n_features, 1)
# Define the loss function
loss_fn = nn.MSELoss(reduction='none')
# Robust loss
(*@\textcolor{ForestGreen}{rho = pt.tensor(.1)}@*)
(*@\textcolor{ForestGreen}{robust\_loss = robustify(loss\_fn, model, rho, X, y)}@*)
# Define the optimizer
(*@\textcolor{BrickRed}{optimizer = optim.AdamW(model.parameters(), lr=.1)}@*)
(*@\textcolor{ForestGreen}{optimizer = robust\_loss.optimizer}@*)  # or use your favorite optimizer
# Training loop
for epoch in range(100):
    avg_loss = 0.
    (*@\textcolor{ForestGreen}{robust\_loss.get\_initial\_guess\_at\_dual(X, y)}@*)
    for batch_x, batch_y in train_loader:
        optimizer.zero_grad()
        (*@\textcolor{BrickRed}{loss = loss\_fn(model(batch\_x), batch\_y))}@*)
        (*@\textcolor{ForestGreen}{loss = robust\_loss(batch\_x, batch\_y)}@*)
        loss.backward()
        optimizer.step()
        avg_loss += loss.detach().item()  
\end{lstlisting}

\section{Numerical aspects}\label{sec:issues}

The generic robustification of \texttt{PyTorch} models through the smoothed objective \eqref{eq:EntWDRO} presents some numerical challenges, before even optimizing the model, indeed:

\vspace{-0.5ex}

\begin{itemize}
\item Given an implementation of the loss function $\obj_\param(\sample)$, the objective \eqref{eq:EntWDRO} is not yet amenable to stochastic first-order optimization, due to the presence of a (continuous) expectation inside the logarithm. We approximate this integral by Monte-Carlo sampling, with a bound on the induced bias.

  \vspace{-0.25ex}
\item The presence of the exponential in \eqref{eq:EntWDRO} makes the objective sharply peaked, which results in a high variance in the gradient estimate. We mitigate this with importance sampling
shifting the samples following the gradient  $\nabla_{\!\sample}\obj_\param(\sample_i)$.

\vspace{-0.25ex}
\item The smoothing depends on some parameters $\sigma$, $\varepsilon$ that have to be tuned.   We provide an automatic calibration for them, based on the problem parameters and the following theoretical guidelines of \citet{azizian2024exact}. \end{itemize}

Once the model loss is robustified, the user has two options: either use the new loss in their own training/optimization loop, or use the optimization procedure provided by the library (relying on adaptive optimizers \citep{cutkosky2023mechanic}). 
Numerical illustrations as well as features and implementation details are available in the online documentation at \url{https://skwdro.readthedocs.io}.

\newpage

\acks{This work has been partially supported by MIAI@Grenoble Alpes (ANR-19-P3IA-0003)}

\vskip 0.2in
\bibliography{references}

@PREAMBLE{ "\def\cprime{$'$} " }

@PREAMBLE{ "\def\cdprime{$''$} " }

@STRING{jmlr = "Journal of Machine Learning Research"}

@STRING{nips = "NeurIPS"}

@STRING{mprog = "Mathematical Programming"}

@article{kuhn2025distributionally,
  title={Distributionally robust optimization},
  author={Kuhn, Daniel and Shafiee, Soroosh and Wiesemann, Wolfram},
  journal={Acta Numerica},
  volume={34},
  pages={579--804},
  year={2025},
  publisher={Cambridge University Press}
}

@article{le2025unregularized,
  title={Unregularized limit of stochastic gradient method for Wasserstein distributionally robust optimization},
  author={Le, Tam},
  journal={arXiv preprint arXiv:2506.04948},
  year={2025}
}

@Article{	  azizian2022regularization,
 author = {Azizian, Wa\"{\i}ss and Iutzeler, Franck and Malick, J\'er\^ome},
 title = {Regularization for {W}asserstein distributionally robust optimization},
 journal = {ESAIM: COCV},
 year = 2023,
 _volume = 29,
 _pages = "33",
}

@InCollection{	  blanchet2021statistical,
  title		= {Statistical analysis of {W}asserstein distributionally
		  robust estimators},
  author	= {Blanchet, Jose and Murthy, Karthyek and Nguyen, Viet Anh},
  booktitle	= {Tutorials in Operations Research: Emerging Optimization
		  Methods and Modeling Techniques with Applications},
  _pages		= {227--254},
  year		= {2021},
  publisher	= {INFORMS}
}

@Article{	  esfahani2018data,
  title		= {Data-driven distributionally robust optimization using the
		  {W}asserstein metric: Performance guarantees and tractable
		  reformulations},
  author	= {Esfahani, Peyman Mohajerin and Kuhn, Daniel},
  journal = mprog,
  _journal	= {Mathematical Programming},
  _volume	= {171},
  _number	= {1},
  _pages		= {115--166},
  year		= {2018},
  publisher	= {Springer}
}

@Book{		  gray,
  title		= {Toeplitz and circulant matrices: A review},
  author	= {Gray, R. M.},
  year		= {2006},
  publisher	= {Now Pub}
}

@InProceedings{	  koh2021wilds,
  title		= {Wilds: A benchmark of in-the-wild distribution shifts},
  author	= {Koh, Pang Wei and Sagawa, Shiori and Marklund, Henrik and
		  Xie, Sang Michael and Zhang, Marvin and Balsubramani,
		  Akshay and Hu, Weihua and Yasunaga, Michihiro and Phillips,
		  Richard Lanas and Gao, Irena and others},
  booktitle	= {International Conference on Machine Learning},
  pages		= {5637--5664},
  year		= {2021},
  organization	= {PMLR}
}

@InCollection{	  kuhn2019wasserstein,
  title		= {Wasserstein distributionally robust optimization: Theory
		  and applications in machine learning},
  author	= {Kuhn, Daniel and Esfahani, Peyman Mohajerin and Nguyen,
		  Viet Anh and Shafieezadeh-Abadeh, Soroosh},
  booktitle	= {Operations Research \& Management Science in the Age of
		  Analytics},
  year		= {2019},
  publisher	= {INFORMS}
}

@Article{	  pedregosa2011scikit,
  title		= {Scikit-learn: Machine learning in Python},
  author	= {Pedregosa, Fabian and Varoquaux, Ga{\"e}l and Gramfort,
		  Alexandre and Michel, Vincent and Thirion, Bertrand and
		  Grisel, Olivier and Blondel, Mathieu and Prettenhofer,
		  Peter and Weiss, Ron and Dubourg, Vincent and others},
  journal = jmlr,
  _journal	= {the Journal of machine Learning research},
  _volume	= {12},
  _pages		= {2825--2830},
  year		= {2011},
  publisher	= {JMLR. org}
}

@article{diamond2016cvxpy,
  author  = {Steven Diamond and Stephen Boyd},
  title   = {{CVXPY}: {A} {P}ython-embedded modeling language for convex optimization},
  journal = {Journal of Machine Learning Research},
  year    = {2016},
  _volume  = {17},
  _number  = {83},
  _pages   = {1--5},
}

@article{blanchet2024distributionally,
  title={Distributionally Robust Optimization and Robust Statistics},
  author={Blanchet, Jose and Li, Jiajin and Lin, Sirui and Zhang, Xuhui},
  journal={arXiv preprint arXiv:2401.14655},
  year={2024}
}

@article{belbasi2023s,
  title={It's All in the Mix: Wasserstein Machine Learning with Mixed Features},
  author={Belbasi, Reza and Selvi, Aras and Wiesemann, Wolfram},
  journal={arXiv preprint arXiv:2312.12230},
  year={2023}
}

@Article{	  peyre2019computational,
  title		= {Computational optimal transport: With applications to data
		  science},
  author	= {Peyr{\'e}, Gabriel and Cuturi, Marco},
  journal	= {Foundations and Trends{\textregistered} in Machine
		  Learning},
  _volume	= {11},
  _number	= {5-6},
  _pages		= {355--607},
  year		= {2019},
  publisher	= {Now Publishers, Inc.}
}

@Article{	  shafieezadehabadehregularizationmasstransportation2019,
  ids		= {shafieezadehabadehRegularizationMassTransportation},
  title		= {Regularization via Mass Transportation},
  author	= {{Shafieezadeh-Abadeh}, Soroosh and Kuhn, Daniel and
		  Esfahani, Peyman Mohajerin},
  year		= {2019},
  journal = jmlr,
  _journal	= {Journal of Machine Learning Research},
  _volume	= {20},
  _pages		= {1--68}
}

@inproceedings{
le2025universal,
title={Universal generalization guarantees for Wasserstein distributionally robust models},
author={Tam Le and Jerome Malick},
booktitle={The Thirteenth International Conference on Learning Representations},
year={2025},
url={https://openreview.net/forum?id=0h6v4SpLCY}
}

@article{wang2021sinkhorn,
author = {Wang, Jie and Gao, Rui and Xie, Yao},
title = {Sinkhorn Distributionally Robust Optimization},
journal = {Operations Research},
pages = {1-23},
year = {2025},
doi = {10.1287/opre.2023.0294},
}

@article{chen2020distributionally,
  title={Distributionally robust learning},
  author={Chen, Ruidi and Paschalidis, Ioannis Ch and others},
  journal={Foundations and Trends{\textregistered} in Optimization},
  _volume={4},
  _number={1-2},
  _pages={1--243},
  year={2020},
  publisher={Now Publishers, Inc.}
}

@article{azizian2024exact,
  title={Exact generalization guarantees for (regularized) wasserstein distributionally robust models},
  author={Azizian, Wa{\"\i}ss and Iutzeler, Franck and Malick, J{\'e}r{\^o}me},
  journal=nips,
  _journal={Advances in Neural Information Processing Systems},
  _volume={36},
  year={2024}
}

@article{cutkosky2023mechanic,
  title={Mechanic: A learning rate tuner},
  author={Cutkosky, Ashok and Defazio, Aaron and Mehta, Harsh},
  journal = nips,
  _journal={Advances in Neural Information Processing Systems},
  _volume={36},
  year={2023}
}

@article{liu2025dro,
  title={DRO: A Python Library for Distributionally Robust Optimization in Machine Learning},
  author={Liu, Jiashuo and Wang, Tianyu and Lam, Henry and Namkoong, Hongseok and Blanchet, Jose},
  journal={arXiv preprint arXiv:2505.23565},
  year={2025}
}

@inproceedings{
mehta2024distributionally,
title={Distributionally Robust Optimization with Bias and Variance Reduction},
author={Ronak Mehta and Vincent Roulet and Krishna Pillutla and Zaid Harchaoui},
booktitle={The Twelfth International Conference on Learning Representations},
year={2024},
url={https://openreview.net/forum?id=TTrzgEZt9s}
}

\appendix

\section{Illustration}

In this appendix, we showcase the ability of \texttt{SkWDRO} to handle non-convex loss function\;$f_\theta$, like neural networks. This is a significative feature of our library compared to existing software. We refer to the online documentation (\url{https://skwdro.readthedocs.io/}) for further material (discussion, comparisons, and illustrations).

We consider an image classification problem with the \texttt{iWildsCam} data set (\cite{koh2021wilds}), pretreated to extract a set of frozen features as per \cite{mehta2024distributionally}. This data noticeably suffers from a distribution shift.
For the objective function \eqref{eq:EntWDRO}, we consider a neural network with one hidden layer of $64$ neurons, equipped with a \texttt{Leaky-ReLU} activation function and cross-entropy loss.
The optimal transport ground cost is the squared euclidean norm on the input features, specified as \texttt{"t-NC-2-2"}. The regularization strength $\varepsilon$ is set to $10^{-3}$, and the noise level $\sigma$ on the reference gaussian distribution is set to $10^{-4}$.

\begin{figure}[ht!]
    \centering
    \includegraphics[width=0.8\linewidth]{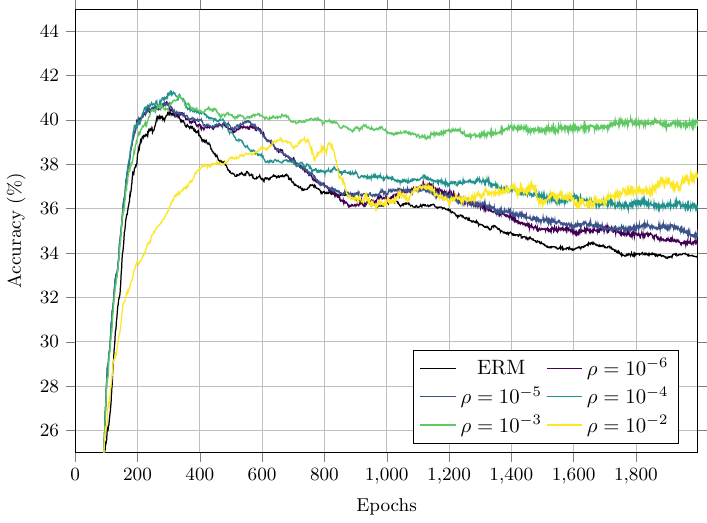}
    \caption{Evolution, over the training epochs, of the test accuracy the neural network on the data set \texttt{iWildsCam}. Colors represent different robustness radii $\rho$.}
    \label{fig:training_wilds}
\end{figure}

Figure~\ref{fig:training_wilds} reports the testing accuracy for a range of robustness radii $\rho\in \{10^{-6},\dots,10^{-1}\}$. We observe that, for small values of $\rho$, the accuracy raises at the beginning, and drops as the training procedure overfits the training set.
In contrast, for higher values of $\rho$, the test performances are superior and do not degrade along training, better defending against the distribution shift.

\end{document}